\documentclass[runningheads]{llncs}
\usepackage{graphicx}
\usepackage{amsmath,amssymb} % define this before the line numbering.
\usepackage{xcolor}
\usepackage[width=122mm,left=12mm,paperwidth=146mm,height=193mm,top=12mm,paperheight=217mm]{geometry}

\usepackage[colorlinks,linkcolor=blue,anchorcolor=green,citecolor=green]{hyperref}
\begin{document}
% \linenumbers
\pagestyle{headings}
\mainmatter
\def\ECCV16SubNumber{***}  % Insert your submission number here

\title{An Improved Baseline Framework for Pose Estimation Challenge at ECCV 2022 Visual Perception for Navigation in Human Environments Workshop} % Replace with your title

% \titlerunning{Summary of A Pose Estimation Framework}

% \authorrunning{Fu et al.}

\author{Jiajun Fu, Yonghao Dang, Ruoqi Yin, Shaojie Zhang, Feng Zhou, Wending Zhao, Jianqin Yin\thanks{Corresponding Author}}
\institute{Beijing University of Posts and Telecommunications, Beijing, China}

\maketitle

\begin{abstract}
This technical report describes our first-place solution to the pose estimation challenge at ECCV 2022 Visual Perception for Navigation in Human Environments Workshop. In this challenge, we aim to estimate human poses from in-the-wild stitched panoramic images. Our method is built based on Faster R-CNN \cite{ren2015faster} for human detection, and HRNet \cite{sun2019deep} for human pose estimation. We describe technical details for the JRDB-Pose dataset, together with some experimental results. In the competition, we achieved 0.303 $\text{OSPA}_{\text{IOU}}$ and 64.047\% $\text{AP}_{\text{0.5}}$ on the test set of JRDB-Pose.
\end{abstract}

%%%%%%%%
\section{Introduction}

Pose estimation plays an important role in computer vision, which serves as a fundamental task for high-level human action analysis, such as action recognition \cite{iqbal2017pose}, action assessment \cite{xiaohan2015joint}, and human-robot interaction \cite{song2012continuous}. In recent years, there has been a growing interest in developing pose estimation models that can accurately detect and estimate human poses in wild scenarios. Although many large-scale datasets \cite{lin2014microsoft,sun2020scalability,doering2022posetrack21} have been introduced in recent years, no dataset primarily targets robotic perception tasks in social navigation environments, and thus rarely reflects the specific challenges found in human-robot interaction and robot navigation in crowded human environments. The JRDB-Pose dataset \cite{vendrow2022jrdb} aims at bridging this gap and sets up a new benchmark for pose estimation.

This technical report focuses on adopting image-based pose estimation, where bottom-up and top-down methods are two commonly used approaches. In bottom-up methods, the pose of individual body parts is first estimated and then grouped to form a complete pose. On the other hand, top-down methods first detect the entire body and then estimate the pose of each body part. A main strength of top-down methods is their ability to incorporate high-level information about the body, such as prior knowledge about human anatomy and shape. This makes them more robust to occlusion and deformation of body parts, which can be a challenge for bottom-up methods. Moreover, panoramic images contain lots of people with different scales. Compared with bottom-up methods, top-down methods are more robust to multi-scale targets. Therefore, we adopt a classical top-down method-HRNet \cite{sun2019deep} for human pose estimation. Details are given in the remainder of this report.

%%%%%%%%
\section{Methods}

We directly adopted Faster R-CNN for human detection and HRNet for pose estimation\footnote{For detailed network architectures, please check the original papers \cite{ren2015faster,sun2019deep}.}. These two networks rely heavily on pre-trained models. For the HRNet, a standard initialization setting is to load pre-trained weights from the COCO dataset \cite{lin2014microsoft}. However, the annotated keypoint locations in the COCO dataset are different from the ones in the JRDB-Pose dataset. To handle this keypoint mismatch, we modified the weights of the last convolution (from the feature space to the keypoint space). For a keypoint in the JRDB-Pose dataset, we find the nearest counterpart(s) in the COCO dataset and copy the last convolution weights for this counterpart. If there are multiple counterparts, we utilize the average weights. We name this nearest-match-based weight initialization. The detailed correspondences are shown in Table \ref{tab:keypoint_ref}.

\begin{table}[!htb]
\setlength\tabcolsep{3pt} % 每一列的宽度
\caption{COCO's counterparts for keypoints in the JRDB-Pose dataset}
\label{tab:keypoint_ref}
\centering
\begin{tabular}{ccc}
\hline
Index & Name & Counterpart Name \\
\hline
1 & head & left eye, right eye \\
2 & right eye & right eye \\
3 & left eye & left eye \\
4 & right shoulder & right shoulder \\
5 & neck & left shoulder, right shoulder \\
6 & left shoulder & left shoulder  \\
7 & right elbow & right elbow \\
8 & left elbow & left elbow \\
9 & center hip & left hip, right hip \\
10 & left hand & left wrist \\
11 & right hip & right hip \\
12 & left hip & left hip \\
13 & left hand & left wrist \\
14 & right knee & right knee \\
15 & left knee & left knee \\
16 & right foot & right ankle \\
17 & left foot & left ankle \\
\hline
\end{tabular}
\end{table}

%%%%%%%%
\section{Experiments}

\subsection{Implementation Details}

\subsubsection{Dataset.} For 2022's challenge of human pose estimation, the JRDB-Pose dataset \cite{vendrow2022jrdb} provides 56,000 stitched panoramic images and 600,000 bounding box annotations, making JRDB-Pose one of the most extensive publicly available datasets that provide ground truth human body pose annotations. Moreover, these annotations come from in-the-wild videos and incorporate heavily occluded poses, which makes JRDB-Pose challenging and an accurate representation of the real-world environment. For the needs of the challenge, we select training images in the "Bytes-coffee", "Huang", "Huang-hall", "Huang-lane", and "Jordan-hall" as the validation set. The rest images are used as the training set. Please refer to the JRDB-Pose paper \cite{vendrow2022jrdb} for the detailed calculation of evaluation metrics.

\subsubsection{Training.} We trained the detection model and pose estimation model, separately. Each model is trained in an end-to-end manner.

For the detection model, we trained the Faster R-CNN \cite{ren2015faster} using ResNet-50 \cite{he2016deep} as the backbone. It adopted SGD with a mini-batch of four on four 3090 GPUs and trained it for 50 epochs with a base learning rate of 0.02, which is decreased by a factor of 10 at epochs 30 and 45. We performed a linear warm-up \cite{goyal2017accurate} during the first 1500 iterations. MMDetection \cite{chen2019mmdetection} was used for implementation. Besides, we directly utilized the stitched panoramic images and extracted ground truth bounding boxes based on corresponding poses. We performed a random left/right shift and removed bounding boxes across the left and right sides of the image. 

For the pose estimation model, we trained the HRNet-w48 \cite{sun2019deep} using the pre-trained model on COCO \cite{lin2014microsoft} as initialization weights. It adopted SGD with a mini-batch of 64 on two 3090 GPUs and trained it for 210 epochs with a base learning rate of 0.0005, which is decreased by a factor of 10 at epochs 170 and 200. We performed a linear warm-up \cite{goyal2017accurate} during the first 500 iterations. MMPose \cite{mmpose2020} was used for implementation. Besides, we cropped the input images based on corresponding poses and resized the images into $384 \times 288$. %TODO: augmentation in mmdet.

\subsubsection{Inference.} During testing, we first extracted bounding boxes from the detector model and performed pose estimation. Specifically, we directly input the stitched panoramic images to the Faster R-CNN model. Afterward, we utilized non-maximal suppression (NMS) to filter overlapping detections. The rest detections were resized into $384 \times 288$ and fed into HRNet-w48 to generate pose estimation results. For each model, we utilize the checkpoint with the best performance on the validation set.

\begin{table}[!htb]
\setlength\tabcolsep{3pt} % 每一列的宽度
\caption{Comparisons of different methods on the JRDB-Pose dataset. The best results are highlighted in \textbf{bold}. Please visit \href{https://jrdb.erc.monash.edu/leaderboards/pose}{the leaderboard website} for detailed information.}
\label{tab:main_results}
\centering
\begin{tabular}{ccc}
\hline
Method & $\text{OSPA}_{\text{IOU}} \downarrow$ & $\text{AP}_{\text{0.5}} \uparrow$ \\
\hline
Baseline1 & 0.459 & 43.338\% \\
\hline
Baseline2 & 0.450 & 43.342\% \\
\hline
Ours      & \textbf{0.303} & \textbf{64.047\%} \\
\hline
\end{tabular}
\end{table}

\subsection{Results.}

\begin{figure}[!htb] %H为当前位置，!htb为忽略美学标准，htbp为浮动图形
  \centering %图片居中
  \includegraphics[width=\textwidth]{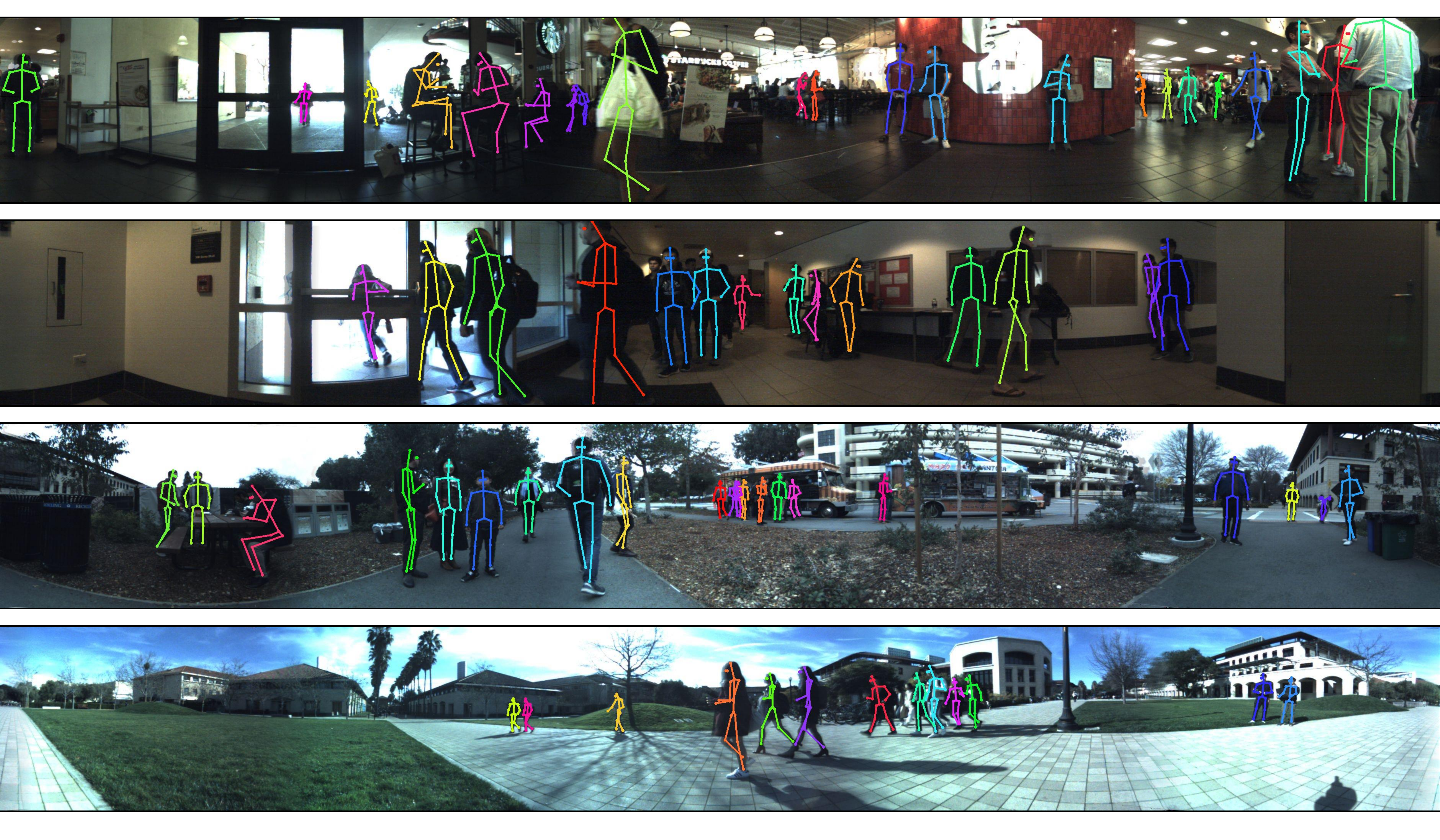} %插入图片，[]中设置图片大小，{}中是图片文件名
  \vspace{-1cm}
  \caption{Qualitative Results of our model on the JRDB-Pose dataset.} %最终文档中希望显示的图片标题
  \label{fig:pose_vis} %用于文内引用的标签
\end{figure}

We compare our model with the state-of-the-art methods on the JRDB-Pose test set. Table \ref{tab:main_results} shows that our framework outperforms existing methods by a large margin. We show some visualization results in Figure \ref{fig:pose_vis}.

\subsection{Ablation Studies}

In this section, we show the effectiveness of the weight initialization and the human detector.

For the weight initialization, we evaluated the adopted initialization in the validation set. Specifically, we tested three initialization and the results are presented in Table \ref{tab:weight_init}. The ``ImageNet" indicates initializing image backbone weights from ImageNet \cite{deng2009imagenet}. This initialization performed the worst since the decoders in HRNet were initialized randomly and get no prior keypoint information. The ``COCO + Random" represents COCO weight initialization except for the last convolution, whose weights were initialized randomly. The performance is slightly worse than our initialization method (``COCO + Match").

For the human detector, we tested bounding boxes from different approaches on the test set. The results are shown in Table \ref{tab:human_detect}. The ``Baseline" indicates pose estimation results from the official bounding boxes. There are many redundant boxes and some bounding boxes are small to cover enough image cues. With the NMS and implemented human detector, the performance improves a lot on the test set.

\begin{minipage}{\textwidth}

\begin{minipage}[t]{0.48\textwidth}
\makeatletter\def\@captype{table}
\caption{Impact of weight initialization}
\label{tab:weight_init}
\setlength\tabcolsep{3pt}
\begin{tabular}{cc}
\hline
Method & $\text{AP}_{\text{0.5}} \uparrow$ \\
\hline
ImageNet & 92.892\% \\
COCO + Random & 96.482\% \\
COCO + Match & 97.968\% \\
\hline
\end{tabular}
\end{minipage}
\begin{minipage}[t]{0.48\textwidth}
\makeatletter\def\@captype{table}
\caption{Impact of human detector}
\label{tab:human_detect}
\setlength\tabcolsep{3pt}
\begin{tabular}{ccc}
\hline
Name & $\text{OSPA}_{\text{IOU}} \downarrow$ & $\text{AP}_{\text{0.5}} \uparrow$ \\
\hline
Baseline & 0.406 & 34.186\% \\
Ours & 0.303 & 64.047\% \\
\hline
\end{tabular}

\end{minipage}
\end{minipage}

%%%%%%%%
\section{Conclusion}
We demonstrate our solution to the pose estimation challenge in this technical report. Despite simple, it has achieved promising results and has the potential for further improvements.

\bibliographystyle{splncs}
\bibliography{main}
\end{document}